\documentclass[11pt]{article}

\usepackage[preprint]{acl}
\usepackage{times}
\usepackage{latexsym}

\usepackage[T1]{fontenc}
\usepackage[utf8]{inputenc}

\usepackage{microtype}

\usepackage{inconsolata}

\usepackage{graphicx}

\usepackage{booktabs}
\usepackage{amsmath}
\usepackage{algorithmic}
\usepackage{algorithm}
\usepackage{amssymb}
\usepackage{subcaption}

\title{One Token Is Enough: Improving Diffusion Language Models \\ with a Sink Token}
\author{
 \textbf{Zihou Zhang\textsuperscript{1}},
 \textbf{Zheyong Xie\textsuperscript{1}},
 \textbf{Li Zhong\textsuperscript{1}},
 \textbf{Haifeng Liu\textsuperscript{1}},
 \textbf{Yao Hu\textsuperscript{1}},
 \textbf{Shaosheng Cao\textsuperscript{1}}
\\
 \textsuperscript{1}Xiaohongshu.inc,
\\
\small{
  \textbf{Code:} \url{https://github.com/skywalker0523/OneTokenIsEnough}
}
}

\begin{document}
\maketitle
\begin{abstract}
Diffusion Language Models (DLMs) have emerged as a compelling alternative to autoregressive approaches, enabling parallel text generation with competitive performance. Despite these advantages, there is a critical instability in DLMs: the moving sink phenomenon. Our analysis indicates that sink tokens exhibit low-norm representations in the Transformer's value space, and that the moving sink phenomenon serves as a protective mechanism in DLMs to prevent excessive information mixing. However, their unpredictable positions across diffusion steps undermine inference robustness. 
To resolve this, we propose a simple but effective extra sink token implemented via a modified attention mask.
Specifically, we introduce a special token constrained to attend solely to itself, while remaining globally visible to all other tokens. 
Experimental results demonstrate that introducing a single extra token stabilizes attention sinks, substantially improving model performance. Crucially, further analysis confirms that the effectiveness of this token is independent of its  position and characterized by negligible semantic  content, validating its role as a robust and dedicated structural sink.
\end{abstract}

\begin{figure*}[t]
  \centering
  \includegraphics[width=\textwidth]{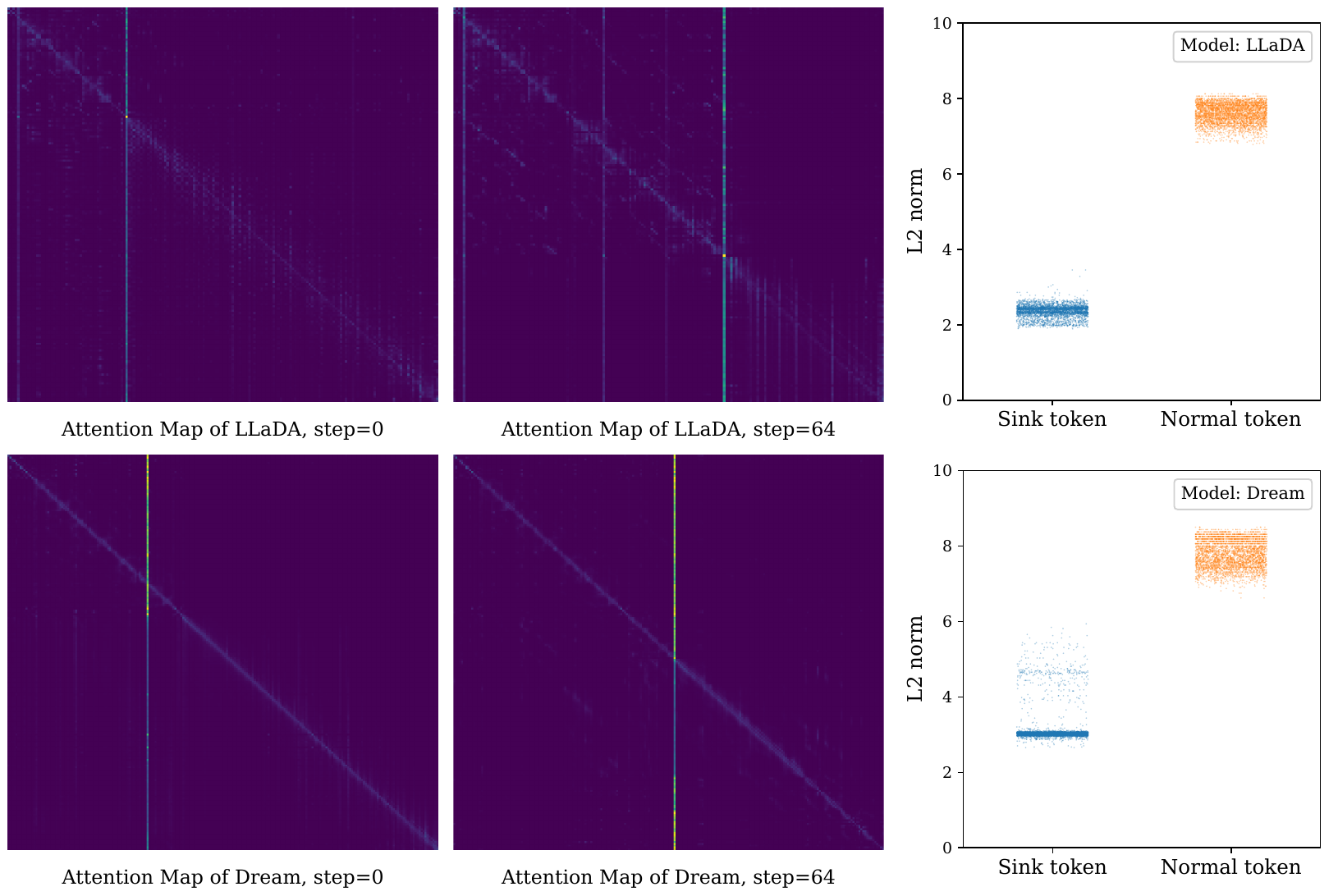}
  \caption{Moving attention sinks and their low-norm property in diffusion language models. Attention maps show representative patterns of LLaDA-Instruct (top) and Dream-Instruct (bottom) at two denoising steps ($step{=}0$ and $step{=}64$). Bright vertical stripes mark sink tokens that attract concentrated attention; their positions shift across steps, demonstrating moving sinks. Norm scatter plots compare the $L_2$ norms of value vectors for 10K sink tokens and 10K normal tokens sampled during inference on GSM8K~\cite{gsm8k} and HumanEval~\cite{humaneval}. Sink tokens exhibit significantly lower mean $L_2$ norms compared to normal tokens (2.36 vs. 7.60 for LLaDA; 3.15 vs. 7.79 for Dream). This consistent disparity suggests that DLMs preferentially select low-norm tokens as sinks during generation.}
  \label{figure_normand_attn}
\end{figure*}

\section{Introduction}

Diffusion Language Models (DLMs) have emerged as a compelling alternative to autoregressive (AR) approaches, introducing a paradigm shift towards parallel text generation. Unlike the left-to-right sequential decoding of AR models, DLMs utilize bidirectional attention to enable global context modeling. This mechanism has demonstrated remarkable potential, with recent works such as LLaDA~\cite{llada} and Dream~\cite{dream} achieving competitive performance in complex generation tasks.

Nevertheless, despite these capabilities, DLMs remain constrained by practical limitations, including inefficient KV caching~\cite{sparse-dllm,fastdllm} and complex optimization strategies~\cite{d1}. Beyond these general issues, a more specific instability arises from the standard Transformer backbone: the \textbf{attention sink} phenomenon. While AR models inadvertently stabilize this phenomenon by anchoring excessive attention to a fixed initial token via causal masking~\cite{streamllm,whysink}, DLMs lack such a structural constraint. In the absence of a consistent start token and causal mask, the sink position in DLMs shifts erratically across diffusion timesteps and layers~\cite{attentionsinkinDLLM}. This unpredictable behavior contrasts sharply with the stability of AR models, posing a unique challenge to inference robustness.

In this work, we analyze the characteristics of sink tokens in DLMs through the lens of the transformer value space. As shown in Figure~\ref{figure_normand_attn}, we observe that sink tokens in DLMs consistently exhibit the lowest $L_2$ norms. This behavior functions as an implicit regularization mechanism, where the model directs attention to these low-norm tokens to mitigate excessive information mixing~\cite{whensink,whysink}, relevant details are shown in Appendix~\ref{appendix:value_space_norm}. Consequently, the inference process inherently offloads excess attention onto tokens with negligible semantic information.

However, this dependency on low-information tokens creates instability due to the stochastic input masking in DLMs. Unlike autoregressive models that utilize a fixed initial token as a stable anchor, DLMs lack a static, information-sparse token. Instead, they typically utilize masked tokens as attention sinks. Since the set of masked tokens changes dynamically across diffusion steps, there is no consistent position for attention offloading. This moving sink phenomenon introduces structural instability during inference and potentially constrains model performance. This raises a pivotal question: \emph{Can we introduce a dedicated, static sink token to effectively regulate information flow and stabilize the moving sink behavior?}

To validate this hypothesis, we introduce an extra position-stable sink token explicitly designed for Diffusion Language Models. This is implemented by prepending the sink token to the beginning of the sequence, utilizing a modified attention mask where the sink token is constrained to attend solely to itself, while remaining globally visible to all other tokens. We demonstrate that this simple architectural modification yields substantial performance improvements across DLMs with various initialization strategies. By providing a stable, dedicated anchor, our method effectively mitigates the instability caused by moving sinks, leading to consistently better generation quality regardless of the underlying model configuration.

Motivated by these significant performance gains, we further explore the mechanism behind this extra sink token through extensive experiments. We first observe that when this low-information token is introduced during pretraining, the model spontaneously learns to offload a significant portion of its attention mass to it. Moreover, our analysis reveals that this effectiveness is position-invariant. Whether the sink token is placed at the beginning or the end of the sequence, the model achieves comparable improvements, confirming that the benefit stems from the token's functional role as a fixed attention sink rather than its specific position in the sequence.

Our main contributions are as follows:
\begin{itemize}
    \item We provide a systematic analysis of \emph{moving sinks} in DLMs from the \emph{value-space} perspective, revealing that sink tokens consistently act as an implicit mechanism to mitigate excessive information mixing.

    \item Based on our hypothesis, we propose a simple yet effective \emph{extra sink token} which not only taming moving sink issue in DLMs but also improves DLMs' performance. Extensive experiments are conducted to prove the effectiveness of our approach.
    
\end{itemize}

\section{Related Work}

\begin{figure*}[t]
  \centering
  \includegraphics[width=\textwidth]{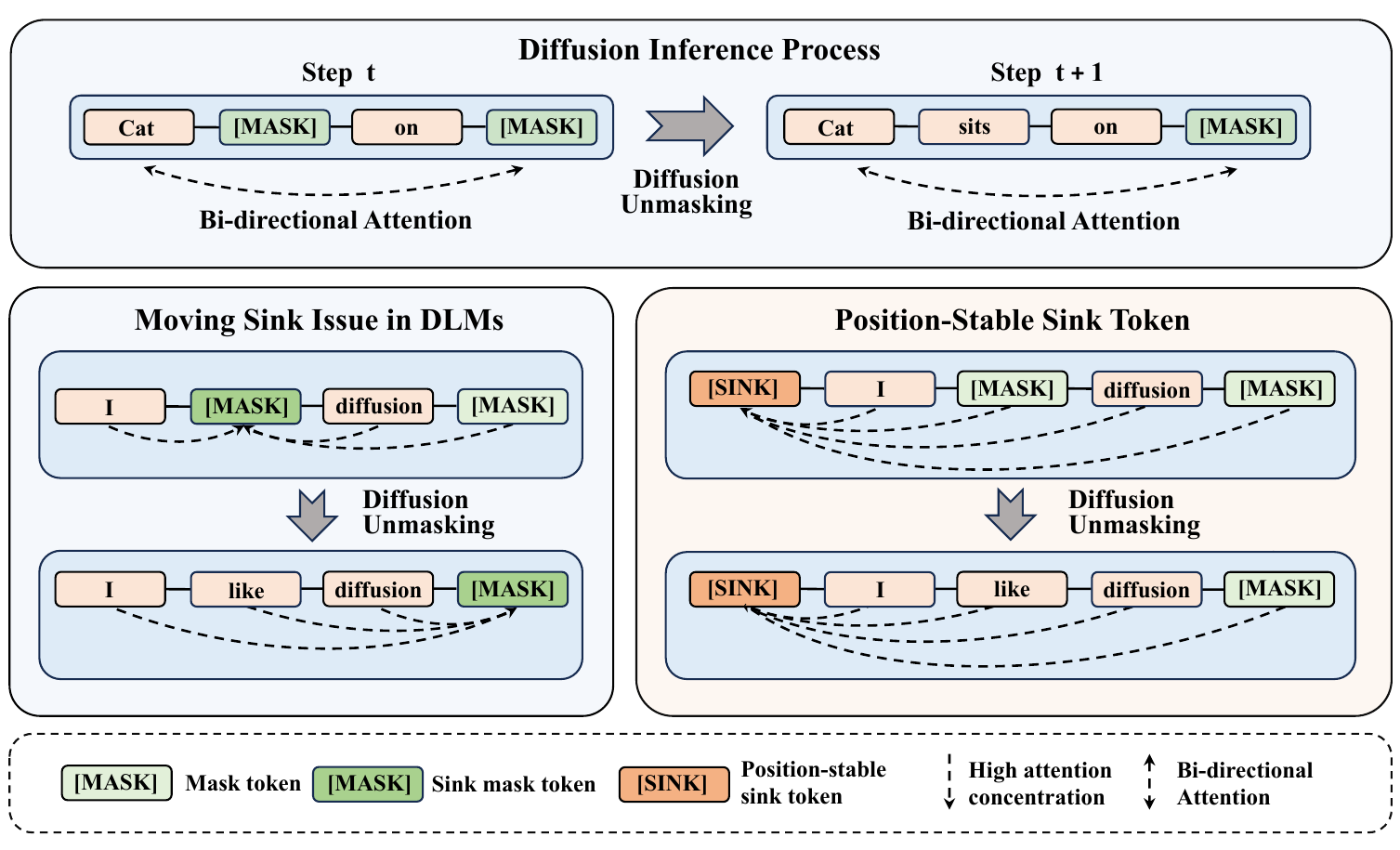}
  \caption{\textbf{Overview of the moving-sink phenomenon in diffusion language models and our stable sink token.}
\textbf{Top:} Diffusion inference iteratively denoises a partially masked sequence; at each step, the model predicts masked positions with bi-directional attention.
\textbf{Middle (left):} In vanilla DLMs, attention can concentrate on low-norm tokens (often a [MASK]) that act as an implicit sink; since the masked set changes across steps, the sink position shifts over time (\emph{moving sink}), which may increase inference instability.
\textbf{Middle (right):} We prepend an extra [SINK] token, turning the moving sink into a stable sink.
\textbf{Bottom:} Legend of the symbols used in the figure.}
  
  \label{figureall}
\end{figure*}

\subsection{Diffusion Language Model}

Diffusion Language Models have been studied from both continuous and discrete perspectives. \citet{likelihooddllm} analyzed scaling laws for continuous diffusion language models and found that, to achieve favorable compute efficiency, continuous diffusion models typically require substantially longer training than autoregressive counterparts. Building on this line of work, recent efforts have scaled DLMs to billions of parameters. \citet{SMDM} demonstrated that 1.1B-scale DLMs are effective for downstream language tasks such as question answering. Rather than training from scratch, \citet{ar2dllm} proposed converting pretrained autoregressive language models into DLMs (DiffuGPT and DiffuLLaMA). In parallel, \citet{llada} introduced LLaDA, an 8B-parameter DLM trained from scratch, achieving performance competitive with LLaMA3-8B~\cite{llama3}. 

Beyond general language modeling, DLMs have been extended to other settings. Mercury Coder~\cite{coder} showed practical applicability in code generation. \citet{dllmtextembedd} explored DLMs for text embedding, and recent studies further investigated post-training techniques for DLMs~\cite{d1,BGPO,llada1_5}. DLMs have also been extended to multimodal modeling~\cite{mmada,lumina,llada-v}.

\subsection{Attention Sink}

\noindent\textbf{Attention sinks in autoregressive Large Language Models:}
Attention sinks refer to tokens that attract a disproportionate amount of attention despite carrying limited semantic content~\cite{streamllm,whysink}. In decoder-only autoregressive LLMs, this often manifests as many heads allocating substantial attention to the first token~\cite{whysink,whensink}. Beyond being an analysis artifact, sinks have been connected to several practical considerations, including streaming/sliding-window attention~\cite{streamllm}, KV-cache efficiency and quantization robustness~\cite{intactkv,adaptkv}. Recent analyses suggest that concentrating attention on a low-information token can help mitigate excessive information mixing in deep and long-context Transformers, and can serve as an approximate no-op for some heads when strong updates are unnecessary~\cite{whysink,whensink}. Empirically, removing or weakening the sink pattern (e.g., dropping <bos> token at inference) can substantially degrade performance, especially in long-context settings~\cite{whysink}. Alternative mechanisms such as gated attention explicitly regulate information flow and can suppress sink behaviors, but introduce additional parameters and architectural complexity~\cite{gatedattention}.

\noindent \textbf{Attention sinks in Diffusion Language Model:} Recent study~\cite{attentionsinkinDLLM} shows that attention sinks also arise in Diffusion Language Models, but with distinct characteristics. Unlike the largely static sinks observed in autoregressive models, sinks in DLMs are often dynamic, shifting across denoising steps due to the bidirectional and iterative nature of diffusion generation. Moreover, sink positions in DLMs are not confined to the beginning of the sequence and frequently align with meaningless tokens such as mask token and white spaces token. Recent work has begun to exploit sink tokens as pivotal tokens to construct sparse attention patterns for accelerating inference process of DLMs~\cite{sparsed,sparse-dllm}.

In this work, we analyze the characteristics of sink tokens in DLMs from the perspective of the Transformer value space. We find that, in DLMs, the moving sink token exhibits the same behavior observed in autoregressive models: it consistently attains the smallest global norm. This phenomenon can be interpreted as a protective mechanism learned during training, which prevents token-wise information from being excessively mixed. We further hypothesize that the position instability of moving sinks introduces additional inference-time instability, potentially constraining further gains in model capacity. Motivated by this hypothesis, we conduct experiments that replace the moving sink with a position-stable sink token, and we observe consistent improvements in performance.

\section{Method}
As shown in Figure~\ref{figureall}, during both training and inference of DLMs, we introduce an additional sink token to convert the unstable moving attention sink into a position-stable attention sink, thereby improving DLM performance.

\subsection{Diffusion Language Modeling}
\label{sec:dlm}

Diffusion Language Models (DLMs) define a distribution over complete token sequences via (i) a forward masking corruption process and (ii) a learned reverse denoising process. Let \(x_0 = (x_0^1,\ldots,x_0^L)\) denote a clean sequence of length \(L\), where \(x_0^i\) is the token at position \(i\). We use a special mask token [MASK]. The forward process yields progressively corrupted variables \(x_{1:T}=\{x_1,\ldots,x_T\}\), and the reverse process reconstructs less corrupted sequences from more corrupted ones. Unlike autoregressive models, DLMs predict masked tokens in parallel at each denoising step.

\subsubsection{Forward Process}
\label{subsec:dlm_forward}
Starting from \(x_0\), the forward process progressively corrupts the sequence by replacing tokens with [MASK], producing \(x_{1:T}\). It factorizes as
\begin{equation}
q(x_{1:T} \mid x_0) = \prod_{t=1}^{T} q(x_t \mid x_{t-1}),
\end{equation}
where \(t\in\{1,\ldots,T\}\) indexes diffusion steps and \(T\) is the total number of steps. Each transition \(q(x_t \mid x_{t-1})\) applies independent masking per position under a predefined noise schedule: at step \(t\), a token remains \emph{unmasked} with some probability, otherwise it is replaced by [MASK]. We set \(x_T\) to be fully corrupted (all positions are [MASK]), so generation can start from a fixed maximally corrupted sequence and then denoise.

\subsubsection{Reverse Process}
\label{subsec:dlm_reverse}
The reverse process iteratively denoises from \(x_T\) back to \(x_0\). We parameterize reverse transitions with a neural model \(p_\theta\) that predicts tokens conditioned on the current corrupted sequence, i.e., \(p_\theta(x_{t-1} \mid x_t)\). Operationally, \(p_\theta\) performs \emph{mask prediction}: it predicts token identities at masked positions conditioned on \(x_t\), and it predicts all masked tokens in parallel. A common choice treats positions independently given \(x_t\), yielding
\begin{equation}
\label{eq:reverse_factorization}
p_\theta(x_0 \mid x_t)
=
\prod_{i=1}^{L} p_\theta(x_0^i \mid x_t).
\end{equation}
During generation, these conditionals are used to fill masked tokens to obtain a less corrupted sequence, repeating over steps until a fully specified \(x_0\) is reached.

\subsubsection{Training Objective}
\label{subsec:dlm_objective}
DLMs are trained with a cross-entropy objective computed only on masked positions.
Let $\{\tau_t\}_{t=1}^{T}$ denote a predefined noise schedule, where $\tau_t\in(0,1]$ is the masking probability at diffusion step $t$.
For a corrupted sequence $x_t$ produced from $x_0$, we define the masked-token log-likelihood as
\begin{equation}
\label{eq:masked_token_loss}
\begin{aligned}
\ell(x_0, x_t; \theta)
&=
\frac{1}{\tau_t}
\sum_{i=1}^{L}
\mathbf{1}\!\left[x_t^{i} = [MASK]\right]
\\
&\qquad
\log p_\theta\!\left(x_0^{i} \mid x_t\right) .
\end{aligned}
\end{equation}
The indicator restricts the loss to masked positions. The overall training objective minimizes the negative expected masked-token log-likelihood:
\begin{equation}
\label{eq:dllm_loss}
\mathcal{L}(\theta)
=
- \mathbb{E}_{t,\,x_0,\,x_t}
\big[
\ell(x_0, x_t; \theta)
\big] .
\end{equation}

This objective learns a denoising function over partially observed sequences, enabling parallel token updates
and avoiding strict left-to-right dependencies.

\subsection{Extra Sink Token for DLM}

As illustrated in Figure~\ref{figure_normand_attn}, moving sink behavior is widely observed in existing DLMs. 
In the absence of a fixed low-norm anchor, the attention sink tends to shift unpredictably across tokens. These temporary sink tokens typically contain low semantic information and exhibit a lower $L_2$ norm. 
The root cause lies in the softmax operation which necessitates that attention weights sum to one, mirroring the attention sink behavior observed in autoregressive models~\cite{whysink}.
When a token lacks a strong semantic match within the context, the model is forced to assign redundant attention mass to globally visible tokens, often turning content tokens into implicit sinks.

However, this shifting behavior complicates the modeling of the diffusion process and hinders the improvement of DLMs. To address this issue and stabilize the attention mechanism, we introduce a dedicated extra sink token to the input sequence. This token is designed to serve as a stable, low-information target to absorb excess attention. To strictly enforce its role as a pure sink and prevent it from aggregating semantic information, we apply a structured attention mask with the following constraints: (1) the sink token is restricted to attend only to itself; and (2) all other tokens in the sequence are allowed to attend to the sink token.

Formally, given an input sequence of latent representation $X \in \mathbb{R}^{L \times d_{\mathrm{model}}}$, we prepend an extra sink token embedding $s \in \mathbb{R}^{d_{\mathrm{model}}}$ to the sequence. 
The augmented input $\tilde{X} \in \mathbb{R}^{(L+1) \times d_{\mathrm{model}}}$ is given by:
\begin{equation} 
\label{addtokninfront} 
\tilde{X} = [s; X]. 
\end{equation}
Let $k=0$ denote the index of the sink token $s$, we enforce the sink constraints by defining the attention mask bias $M_{ij}$ as: 
 \begin{equation} 
 \label{eq:sink_mask} 
 M_{ij} = \begin{cases} -\infty, & \text{if } i = k \text{ and } j \neq k \\ 
 0, & \text{otherwise} 
 \end{cases} 
 \end{equation}
Under this formulation, the sink token is effectively isolated from aggregating sequence information, creating an asymmetric dependency where content tokens retain the ability to allocate attention mass to it. This constraint ensures that the sink token remains semantically neutral, functioning solely as a target to absorb excess attention weights. In our DLM framework, this configuration is applied consistently across all diffusion timesteps. Given that it introduces only a single additional token and utilizes standard masking, the computational overhead is negligible, providing an efficient and effective solution to regulate attention behavior and enhance model robustness.

\section{Experiment}

\begin{table*}[t]
\centering
\small
\setlength{\tabcolsep}{6pt}
\begin{tabular}{lccccccccc}
\toprule
\textbf{Model} & \textbf{Tokens} & \textbf{ARC-e} & \textbf{ARC-c} & \textbf{HellaSwag} & \textbf{PIQA} & \textbf{RACE} & \textbf{SIQA} & \textbf{LAMBADA} & \textbf{GSM8K} \\
\midrule
\multicolumn{10}{c}{\textbf{0.5B Models}} \\
\cmidrule(lr){1-10}
DLM          & 30B  & 44.70 & 27.13 & 35.21 & 55.28 & 31.29 & \textbf{40.28} & 44.09 & 46.92 \\
DLM + extra token     & 30B  & \textbf{55.47} & \textbf{32.17} & \textbf{48.91} & \textbf{63.44} & 33.79 & 37.82 & \textbf{45.24} & \textbf{49.20} \\
DLM + GA     & 30B  & 47.85 & 29.27 & 41.95 & 61.92 & \textbf{34.07} & 36.08 & 45.10 & 47.01 \\
\midrule
\multicolumn{10}{c}{\textbf{1.5B Models}} \\
\cmidrule(lr){1-10}
DLM           & 100B & 65.49 & 41.47 & 59.50 & 66.49 & \textbf{37.80} & 36.18 & \textbf{66.58} & 57.77 \\
DLM + extra token     & 100B & \textbf{68.18} & \textbf{43.43} & \textbf{61.31} & \textbf{68.17} & 37.61 & \textbf{39.97} & 66.41 & \textbf{58.45} \\
DLM + GA     & 100B & 53.03 & 32.27 & 52.90 & 60.33 & 37.42 & 38.26 & 51.10 & 52.35 \\
\bottomrule
\end{tabular}
\caption{Evaluation of our DLMs (autoregressive LLMs initialized). We report results for two model scales (0.5B and 1.5B), separated into two blocks for clarity. The ``Tokens'' column denotes the total number of training tokens used for each setting. ``DLM + extra token'' denotes the DLM augmented with an additional introduced sink token, while ``DLM + GA'' denotes the DLM equipped with the gated attention (GA) mechanism.}
\label{tab:armdm_all}
\end{table*}

\begin{table*}[t]
\centering
\small
\setlength{\tabcolsep}{6pt}
\begin{tabular}{lccccccccc}
\toprule
\textbf{Model} & \textbf{Tokens} & \textbf{ARC-e} & \textbf{ARC-c} & \textbf{HellaSwag} & \textbf{PIQA} & \textbf{RACE} & \textbf{SIQA} & \textbf{LAMBADA} &\textbf{GSM8K}\\
\midrule
 DLM      & 100B   & 32.20 & \textbf{25.09} & 31.56 & 54.08 & 31.77 & 35.98 &   29.26& 34.95\\
 DLM + extra token  & 100B  & \textbf{40.91} & 22.35 & \textbf{40.43} & \textbf{62.24} & 32.73 & \textbf{37.92} & \textbf{47.16}& 35.17 \\
 DLM + GA  & 100B  & 36.28 & 23.46 & 37.61 & 57.07 & \textbf{33.68} & 34.44 & 42.27& \textbf{36.39}\\

\bottomrule
\end{tabular}
\caption{Evaluation of DLMs (trained from scratch) on different benchmarks. The ``Tokens'' column denotes the total number of training tokens used for each setting. ``DLM + extra token'' denotes the DLM augmented with an additional introduced sink token, while ``DLM + GA'' denotes the DLM equipped with the gated attention (GA) mechanism. 
\label{tab:mdmorigin}
}
\end{table*}

\subsection{Experiment Settings}
% Test Dataset
\subsubsection{Benchmarks} To enable a thorough evaluation, we evaluate DLMs on widely adopted benchmarks spanning commonsense reasoning and reading comprehension:
HellaSwag ~\cite{hellaswag},
ARC-e ~\cite{arc-eandarc-c}, ARC-c ~\cite{arc-eandarc-c},
PIQA ~\cite{piqa},
SIQA ~\cite{siqa},
RACE ~\cite{race},
and LAMBADA ~\cite{lambada}.
Following prior studies~\cite{SMDM}, we also evaluate mathematical reasoning ability of DLMs on GSM8K~\cite{gsm8k} after the supervised fine-tune (SFT) process.\\

\subsubsection{Implementation Details}
We conduct experiments under two training settings: (1) training a DLM initialized from an autoregressive LLM, and (2) training a DLM from scratch.
For the former, we initialize the DLM with Qwen2.5 Base model weights~\cite{qwen2_5} and consider two model scales, 0.5B~\footnote{https://huggingface.co/Qwen/Qwen2.5-0.5B} and 1.5B~\footnote{https://huggingface.co/Qwen/Qwen2.5-1.5B} parameters.
These models are trained on the FineWeb dataset~\cite{fineweb}.
For the from-scratch setting, we adopt the same model architecture as the SMDM framework~\cite{SMDM}.
In this case, the model has 0.5B parameters and is trained on the SlimPajama dataset~\cite{slimpajama}.
During supervised fine-tuning (SFT), we fine-tune the DLM for 10 epochs on the augmented training data~\cite{gsm8ktrain}, with a context length of 2048.
Additional training details are provided in Appendix~\ref{appendix: Trainingdetails}.

\subsection{Experiment Results}

%\textbf{Comparison with different attention strategies:} 
Under the setting of training Diffusion Language Models from autoregressive LLMs, we compare three DLM modeling strategies at the 0.5B and 1.5B parameter scales: (1) the vanilla DLM, (2) DLM equipped with element-wise gated attention~\cite{gatedattention}, and (3) DLM with an additional sink token. The experimental results are reported in Table~\ref{tab:armdm_all}. Implementation details for gated attention are provided in Appendix~\ref{appendix: gatedattention}.

At the 0.5B scale, both gated attention and the introduction of an additional sink token lead to consistent performance improvements over the vanilla DLM, demonstrating the effectiveness of these modifications. However, at the 1.5B scale, the DLM with gated attention exhibits degraded performance compared to the vanilla baseline. This suggests that, when training DLMs from larger autoregressive LLMs, forcibly introducing additional parameters to implement attention gating may harm model stability. In contrast, our approach of adding an extra sink token continues to yield performance gains at the 1.5B scale, highlighting its robustness and effectiveness under larger model settings.

To further substantiate the effectiveness of our method, we extend our evaluation to training 0.5B-scale diffusion language models from scratch, maintaining the same baseline configurations as in the previous experiments. The results presented in Table~\ref{tab:mdmorigin} demonstrate that our approach yields consistent improvements across different training paradigms, highlighting its robustness independent of the initialization setting.

\subsection{Ablation Studies}
In this part, we conduct ablation studies to analyze the contributions of different components. 
For efficiency, we utilize the Qwen2.5-0.5B checkpoint as the initialization and train on a subset of 30B tokens from FineWeb. 
Based on this setup, we examine the impact of sink token placement and quantity, followed by an in-depth analysis of how these tokens influence attention allocation.

\subsubsection{The position of sink token}

\begin{table}[t]
\centering
\small
\setlength{\tabcolsep}{6pt}
\begin{tabular}{lcc}
\toprule
\textbf{Benchmark} & \textbf{Front} & \textbf{End} \\
\midrule
ARC-e       & 55.47 & \textbf{56.14} \\
ARC-c   & 32.17 & \textbf{32.51} \\
RACE  & \textbf{33.79} & 31.77 \\
HellaSwag  & \textbf{48.91} & 48.16 \\
PIQA  & 63.44 & \textbf{64.31} \\
% \midrule
% Llama-2 (7B)      & 74.49 & 77.68 \\
\bottomrule
\end{tabular}
\caption{Evaluation of different sink token positions.}
\label{tab:sinkposition}
\end{table}

Unlike autoregressive language models where the initial token naturally serves as an attention sink due to causal masking, diffusion language models utilize bidirectional attention with global context visibility. This distinction raises a critical question for DLMs: does the explicit sink token need to be placed at a specific position to function effectively, or is its efficacy position-agnostic? To investigate this, we conduct ablation studies by varying the placement of the sink token within the input sequence, comparing two distinct configurations: prepending it at the beginning versus appending it at the end.

Formally, for the configuration where the sink token is placed at the beginning of the sequence, given an input sequence \( X \in \mathbb{R}^{n \times d_{\mathrm{model}}} \), we prepend a sink token \( s \in \mathbb{R}^{d_{\mathrm{model}}} \) to obtain the final input sequence, as defined in Equation~\ref{addtokninfront}.
For the configuration where the sink token is placed at the end of the sequence, given the same input sequence \( X \in \mathbb{R}^{n \times d_{\mathrm{model}}} \), we append the extra vector \( s \in \mathbb{R}^{d_{\mathrm{model}}} \) to form the final input, which can be expressed as:
\begin{equation}
\label{addtokninback}
\tilde{X} = [X; s].
\end{equation}
The sink token is added at the corresponding position during both training and inference.

The results, reported in Table~\ref{tab:sinkposition}, indicate stable improvements across both configurations. This confirms that our method is robust to positional variations. It further validates that the sink token functions globally as an attention attractor due to its properties, rather than relying on the positional bias typically seen in causal models.

\begin{figure*}[t]
  \centering
  % Put subfigure captions (titles) on top
  \captionsetup[subfigure]{position=top,justification=centering,skip=2pt}

  \begin{subfigure}[t]{0.49\textwidth}
    \centering
    
    \includegraphics[width=\textwidth]{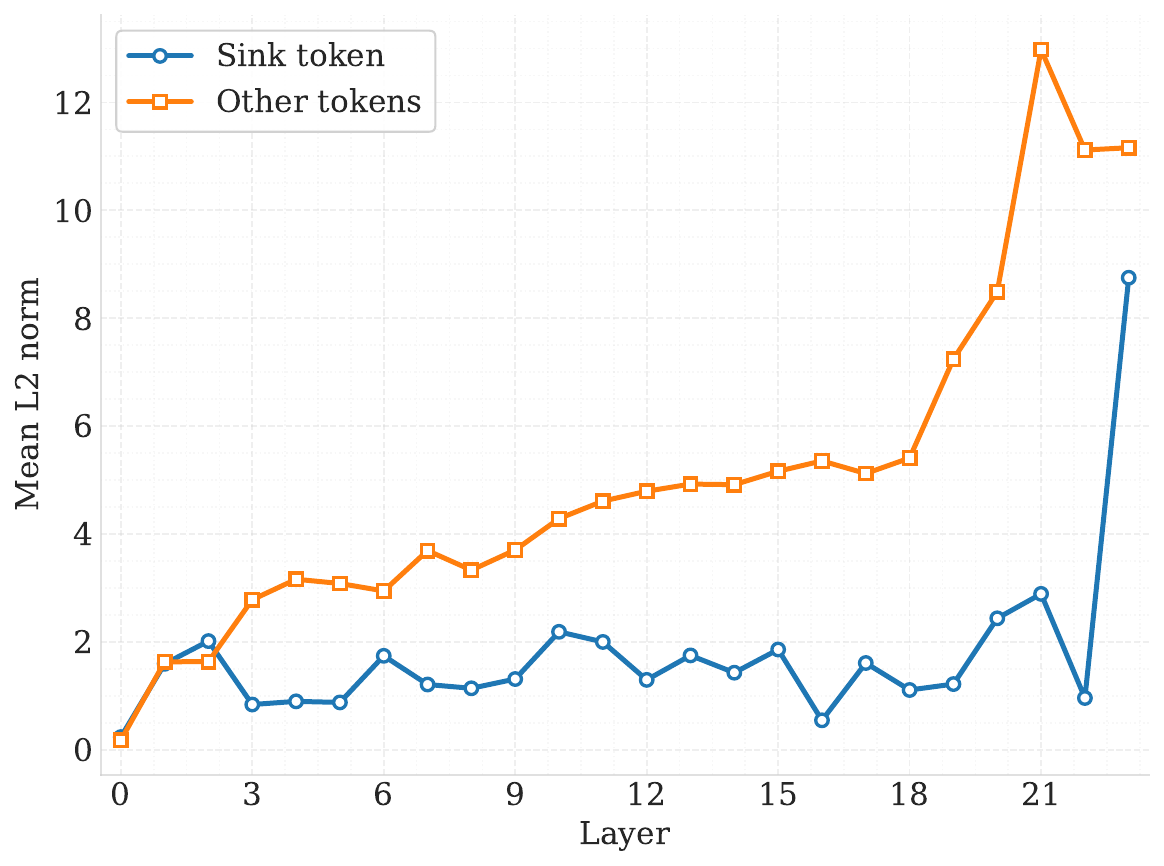}
    \caption{Mean $L_2$ Norm per Layer: Sink vs Others (Model: 0.5B)}
    \label{analyselfmodel:a}
  \end{subfigure}\hfill
  \begin{subfigure}[t]{0.49\textwidth}
    \centering
    
    \includegraphics[width=\textwidth]{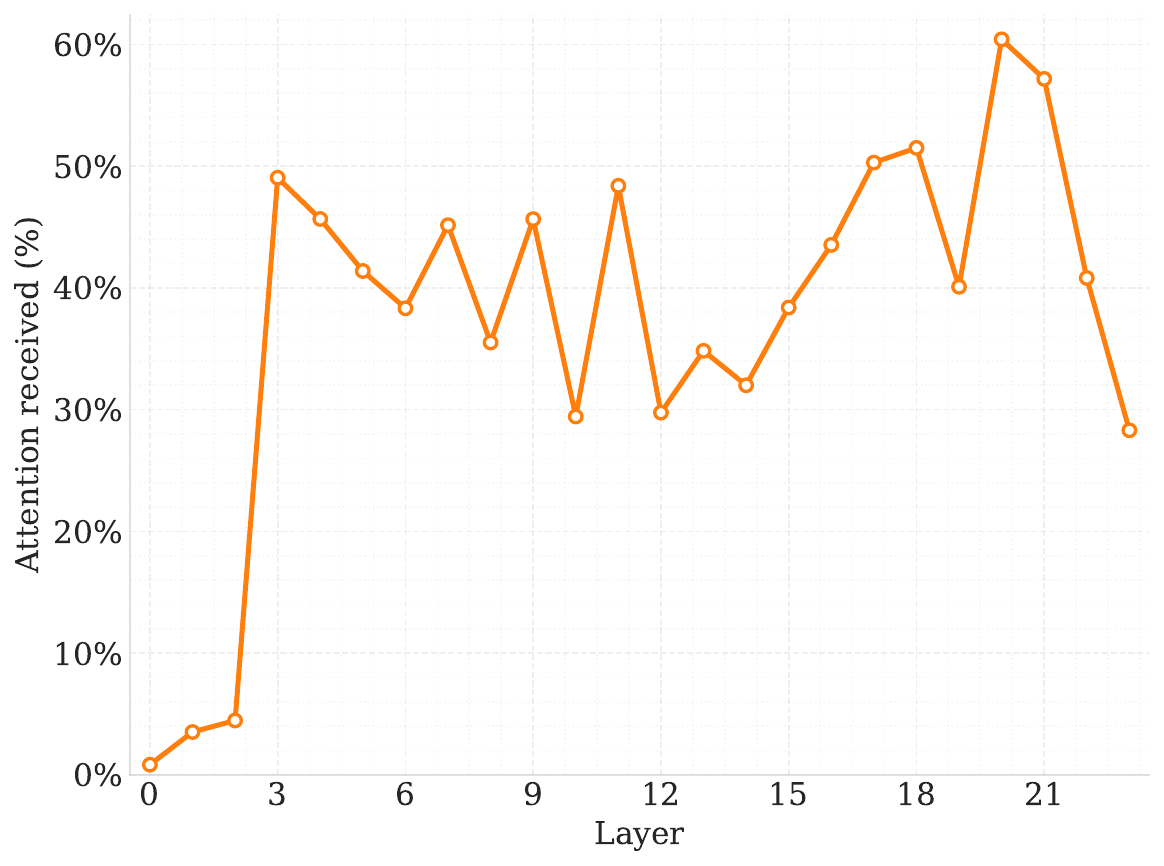}
    \caption{Sink Token Attention Received per Layer (Model: 0.5B)}
    \label{analyselfmodel:b}
  \end{subfigure}

  \vspace{0.6em}

  \begin{subfigure}[t]{0.49\textwidth}
    \centering
    
    \includegraphics[width=\textwidth]{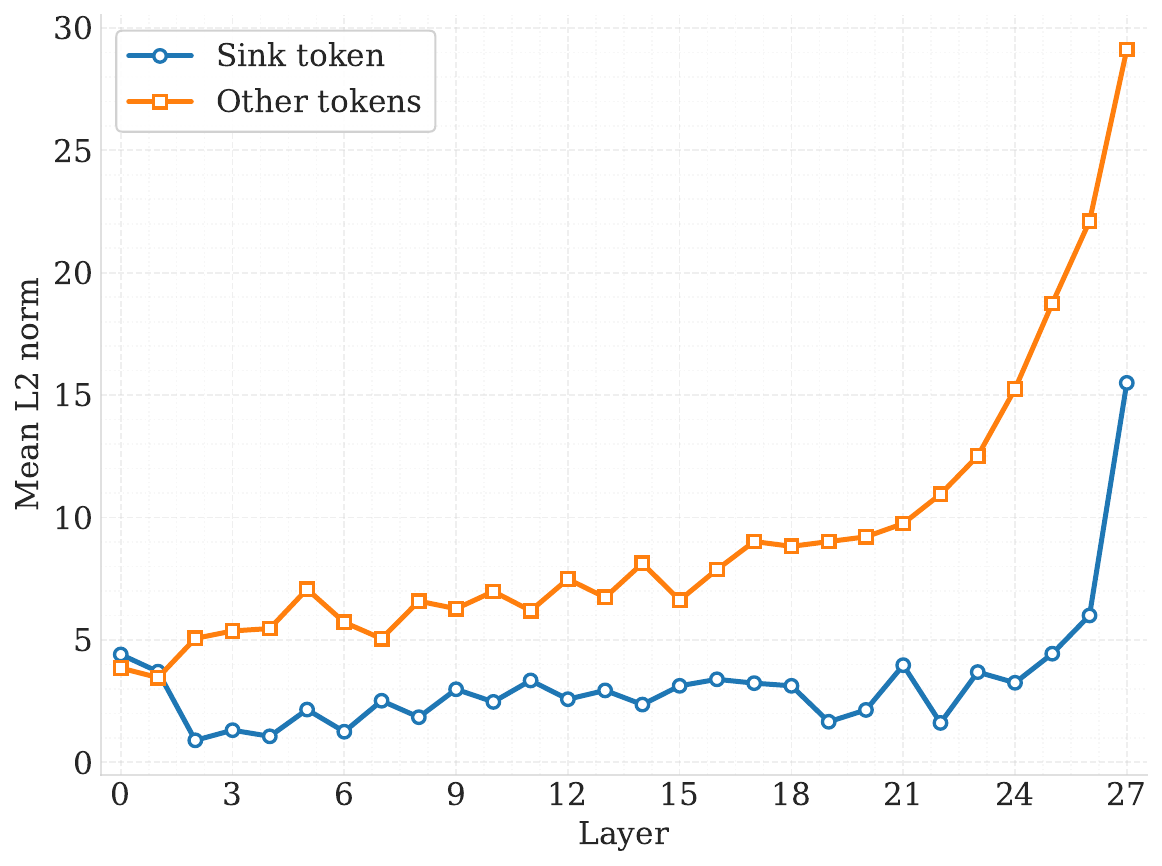}
    \caption{Mean $L_2$ Norm per Layer: Sink vs Others (Model: 1.5B)}
    \label{analyselfmodel:c}
  \end{subfigure}\hfill
  \begin{subfigure}[t]{0.49\textwidth}
    \centering
    
    \includegraphics[width=\textwidth]{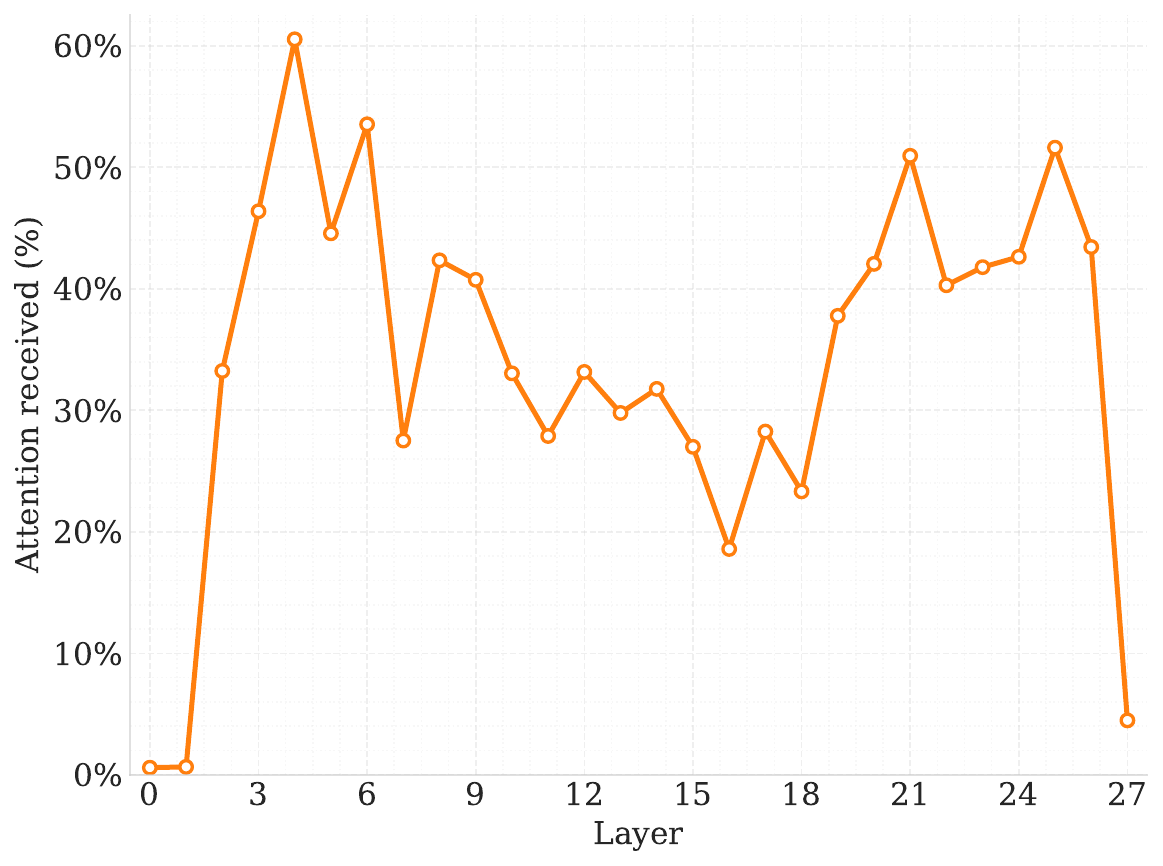}
    \caption{Sink Token Attention Received per Layer (Model: 1.5B)}
    \label{analyselfmodel:d}
  \end{subfigure}

  \caption{Sink token analysis across transformer layers for DLMs initialized from Qwen2.5-0.5B and Qwen2.5-1.5B.
  (a) and (c): mean value-space $L_2$ norm per transformer layer, comparing the extra sink token against all other tokens.
  (b) and (d): the proportion of attention mass received by the sink token per layer.
  After the initial few layers and before the final layers, across the intermediate transformer layers that are most closely related to information integration and transformation, the sink token exhibits substantially smaller value-space $L_2$ norms than other tokens while receiving a large share of global attention.}
  \label{analyselfmodel}
\end{figure*}

\subsubsection{The number of sink token}

%%%%numtoken
\begin{table}[t]
\centering
\small
\setlength{\tabcolsep}{6pt}
\begin{tabular}{lcccc}
\toprule
\textbf{Benchmark} & \textbf{0} & \textbf{1} & \textbf{2} & \textbf{4} \\
\midrule
ARC-e   & 44.70 & \textbf{55.47} & 54.76 & 55.39 \\
ARC-c   & 27.13 & 32.17 & \textbf{32.42} & 31.47 \\
RACE  & 31.29 & \textbf{33.79} & 32.89 & 33.06\\
HellaSwag  & 35.21 & 48.91 & \textbf{48.94} & 48.45 \\
PIQA  & 55.28 & \textbf{63.44} & 63.66 & 62.62\\
% \midrule
% Llama-2 (7B)      & 74.49 & 77.68 \\
\bottomrule
\end{tabular}
\caption{Performance sensitivity to the number of reintroduced tokens on our approach.}
%%%%evaluation of different 
\label{tab:numtoken}
\end{table}

Next, we investigate the impact of sink token quantity to determine if the mechanism's effectiveness is constrained by the capacity of a single token. To this end, we conduct an ablation study by varying the number of added tokens, specifically comparing configurations with 1, 2, and 4 sink tokens. The results, summarized in Table~\ref{tab:numtoken}, reveal that adding a single sink token yields the most substantial improvement, while further increasing the quantity leads to negligible marginal gains. This saturation phenomenon indicates that the sink token functions purely as a structural anchor for attention offloading rather than a carrier of semantic information. Consequently, a single token provides sufficient capacity to fulfill this role.

\subsection{Internal Analysis of Sink Tokens and Attention Allocation}

%%%%sink anneal
\begin{table}[t]
\centering
\small
\setlength{\tabcolsep}{6pt}
\begin{tabular}{lcc}
\toprule
\textbf{Benchmark} & \textbf{Stable sink token} & \textbf{Zero-value token} \\
\midrule
ARC-e       & 55.47 & \textbf{55.68} \\
ARC-c   & 32.17 &  \textbf{32.94} \\
RACE  & \textbf{33.79} & 32.87 \\
HellaSwag  & \textbf{48.91} & 48.81 \\
PIQA  & 63.44 & \textbf{64.09} \\
% \midrule
% Llama-2 (7B)      & 74.49 & 77.68 \\
\bottomrule
\end{tabular}
\caption{Evaluation for setting sink token to zero-vector in value space.}
\label{tab:zerosink}
\end{table}

To validate the effectiveness of our proposed theory beyond standard benchmarks, we conduct a statistical analysis of internal token representations and attention behaviors. Specifically, we examine DLMs initialized from Qwen2.5-0.5B and Qwen2.5-1.5B. Following supervised fine-tuning, we sample token representations in the value space and their corresponding attention maps across Transformer layers during inference on the GSM8K dataset. For each model, we sample 100K inference steps and quantify (i) the \(L_2\) norm statistics of token representations in the value space and (ii) the attention mass allocated to sink tokens.

As illustrated in Figure~\ref{analyselfmodel}, for each layer, we measure the proportion of global attention directed toward the introduced sink token and analyze the distribution of token $L_2$  norms. The results reveal that the sink token maintains a significantly lower $L_2$  norm compared to the average of other tokens while absorbing redundant attention. This indicates that the model implicitly learns to utilize a token with minimal magnitude to offload redundant attention, thereby minimizing interference with the semantic information in the residual stream.

Based on the finding that DLMs benefit from directing attention to tokens with minimal norms, we conduct a further validation study where we explicitly force the sink token’s value states to be zero vectors at every Transformer layer under the setting of training DLM from qwen2.5-0.5B. The results are reported in Table~\ref{tab:zerosink}. We find that even when the sink token is set to a zero vector and contains no semantic information, it still yields consistent gains for the DLM. This confirms our hypothesis that the presence of a low-norm state is the critical factor for mitigating excessive information mixing.

\section{Conclusion}

In this work, we identify the critical role of attention sinks in DLMs to alleviate information over-mixing, while highlighting the positional shifts arising from their distinct attention mechanisms distinct from auto-regressive models. To resolve this, we introduce a position-stable sink token to anchor attention, yielding consistent performance gains across diverse settings. 
Beyond these improvements, we hope our work offers valuable insights into DLM attention mechanics.

\section*{Limitations}
We conducted extensive experiments on Diffusion Language Models at the 0.5B and 1.5B parameter scales. However, we did not scale our experiments to larger Diffusion Language Models.

\section*{Ethics Statement}
We trained and evaluated our model using publicly available datasets that have been verified and widely used. During the preparation of this manuscript, we used generative AI tools for grammar and language proofreading.

\bibliography{custom}

\appendix

\section{Appendix}
\label{sec:appendix}

\subsection{Training details}
\label{appendix: Trainingdetails}

\textbf{Training DLM from autoregressive LLM:} We optimize the model using AdamW~\cite{AdamW}, with \( \beta_1 = 0.9 \), \( \beta_2 = 0.95 \), and a weight decay of \(0.1\).
We adopt a cosine learning-rate schedule, with the peak learning rate set to \(1 \times 10^{-4}\) and the minimum learning rate set to \(1 \times 10^{-5}\), and use linear warmup for the first \(1\%\) of the training tokens.
For the 0.5B model, we train on 30B tokens from the FineWeb~\cite{fineweb} dataset with a batch size of \(512\).
For the 1.5B model, we train on 100B tokens from the FineWeb dataset with a batch size of \(4096\).\\

\textbf{Training DLM from scratch:} To be consistent with the SMDM~\cite{SMDM}, we utilize the AdamW optimizer~\cite{AdamW}, setting $\beta_1 = 0.9$, $\beta_2 = 0.95$, and a weight decay
of $0.1$. Additionally, we apply a cosine learning rate schedule with a maximum
learning rate of $2 \times 10^{-4}$ and a minimum learning rate of
$2 \times 10^{-5}$ with $1\%$ of the tokens for linear warmup. We train the model on 100B tokens from the SlimPajama dataset~\cite{slimpajama}, using a batch size of \(256\).

\subsection{Gated Attention for DLM}
\label{appendix: gatedattention}
We augment the standard attention layer in DLMs with a gating mechanism applied after the scaled dot-product attention (SDPA). The overall Transformer architecture follows the same design as the Qwen model, and the gated attention layer is used as a drop-in replacement for the vanilla attention layer at every diffusion step.

Given the SDPA output
\begin{equation}
\begin{aligned}
Y &= \mathrm{Attention}(Q, K, V) \\
  &= \mathrm{softmax}\!\left(\frac{QK^\top}{\sqrt{d_k}}\right)V.
\end{aligned}
\end{equation}

we apply an element-wise, input-dependent gate to modulate the attention output:

\begin{equation}
Y' = g(Y, X) = Y \odot \sigma(X W_g).
\end{equation}

where $X \in \mathbb{R}^{n \times d_{\mathrm{model}}}$ is the input hidden representation to the attention layer, $W_g \in \mathbb{R}^{d_{\mathrm{model}} \times d_k}$ is a learnable gating projection, $\sigma(\cdot)$ denotes the sigmoid function, and $\odot$ represents element-wise multiplication.

The gated attention output is then passed to the output projection layer:
\begin{equation}
O = Y' W_O,
\end{equation}
where $W_O \in \mathbb{R}^{d_k \times d_{\mathrm{model}}}$.

The gating function introduces non-linearity into the attention layer by dynamically scaling the SDPA output based on the current hidden states. Since the gate is applied after attention weight computation, it preserves the original attention score normalization while enabling adaptive suppression or amplification of attended features. In the context of DLMs, the same gated attention layer is applied across diffusion timesteps, allowing the model to regulate information flow under varying noise conditions with minimal additional computational overhead.

\subsection{Value-Space Token Vectors Analysis}
\label{appendix:value_space_norm}

\subsubsection{Value-space token vectors in a Transformer}
Consider a Transformer layer $\ell$ with input hidden states
$\mathbf{H}^{(\ell)}=[\mathbf{h}^{(\ell)}_1,\ldots,\mathbf{h}^{(\ell)}_n]^\top \in \mathbb{R}^{n\times d}$,
where $n$ is the sequence length and $d$ is the hidden size.
For each attention head $m\in\{1,\ldots,M\}$, the layer computes query, key, and value projections:
\begin{align}
\mathbf{Q}_m &= \mathbf{H}^{(\ell)} \mathbf{W}^{Q}_m, \label{eq:q_def} \\
\mathbf{K}_m &= \mathbf{H}^{(\ell)} \mathbf{W}^{K}_m, \label{eq:k_def} \\
\mathbf{V}_m &= \mathbf{H}^{(\ell)} \mathbf{W}^{V}_m, \label{eq:v_def}
\end{align}
where $\mathbf{W}^{Q}_m,\mathbf{W}^{K}_m,\mathbf{W}^{V}_m \in \mathbb{R}^{d\times d_h}$ and $d_h=d/M$.
The \emph{value-space token vector} of token $j$ (in head $m$) is the $j$-th row of $\mathbf{V}_m$:
\begin{equation}
\mathbf{v}^{(\ell,m)}_j = \mathbf{h}^{(\ell)}_j \mathbf{W}^{V}_m \in \mathbb{R}^{d_h}.
\label{eq:value_vector}
\end{equation}
Intuitively, $\mathbf{v}^{(\ell,m)}_j$ is the content that token $j$ contributes to other tokens through attention mixing.

\subsubsection{$L_2$ norm of a token in value space}
For a head-specific value vector, the $L_2$ norm is computed as
\begin{equation}
\left\|\mathbf{v}^{(\ell,m)}_j\right\|_2
= \sqrt{\sum_{k=1}^{d_h}\left(\mathbf{v}^{(\ell,m)}_{j,k}\right)^2 }.
\label{eq:l2_norm_head}
\end{equation}
To obtain a single magnitude per token across heads, we average per-head norms:
\begin{equation}
\bar{r}^{(\ell)}_j = \frac{1}{M}\sum_{m=1}^{M}\left\|\mathbf{v}^{(\ell,m)}_j\right\|_2.
\label{eq:avg_norm}
\end{equation}
% or concatenate all head values and compute a single norm:
% \begin{align}
% \mathbf{v}^{(\ell)}_j &= \big[\mathbf{v}^{(\ell,1)}_j;\ldots;\mathbf{v}^{(\ell,M)}_j\big]\in\mathbb{R}^{d}, 
% \label{eq:concat_value}
% \end{align}

% \begin{align}
% \left\|\mathbf{v}^{(\ell)}_j\right\|_2 &= \sqrt{\sum_{k=1}^{d}\left(\mathbf{v}^{(\ell)}_{j,k}\right)^2 }.
% \label{eq:l2_norm_concat}
% \end{align}

\subsubsection{Why attending to low-norm tokens approximates a ``no-op''}
In head $m$, the attention output for query token $i$ is a convex combination of value vectors:
\begin{equation}
\mathbf{o}^{(\ell,m)}_i = \sum_{j=1}^{n} \alpha^{(\ell,m)}_{ij}\,\mathbf{v}^{(\ell,m)}_j,
\label{eq:attn_output}
\end{equation}
where the attention weights satisfy
\begin{align}
\sum_{j=1}^{n}\alpha^{(\ell,m)}_{ij} &= 1, \label{eq:attn_sum1}\\
\alpha^{(\ell,m)}_{ij} &\ge 0. \label{eq:attn_nonneg}
\end{align}

% \begin{align}
% % s_i(\theta)=\exp\left(\frac{1}{|y_i|}\sum_{t=1}^{|y_i|}\log\frac{\pi_\theta\!\left(y_{i,t}\mid x,\,y_{i,<t}\right)}{\pi_{\theta_{\mathrm{old}}}\!\left(y_{i,t}\mid x,\,y_{i,<t}\right)}\right).
% \[
% \begin{aligned}
% s_i(\theta)
% &=\exp\!\left(\frac{1}{|y_i|}\sum_{t=1}^{|y_i|}\log\frac{\pi_\theta(y_{i,t}\mid x,y_{i,<t})}{\pi_{\theta_{\mathrm{old}}}(y_{i,t}\mid x,y_{i,<t})}\right)=\exp\!\left(\frac{1}{|y_i|}\log\prod_{t=1}^{|y_i|}\frac{\pi_\theta(y_{i,t}\mid x,y_{i,<t})}{\pi_{\theta_{\mathrm{old}}}(y_{i,t}\mid x,y_{i,<t})}\right)\\
% &=\left(\frac{\prod_{t=1}^{|y_i|}\pi_\theta(y_{i,t}\mid x,y_{i,<t})}{\prod_{t=1}^{|y_i|}\pi_{\theta_{\mathrm{old}}}(y_{i,t}\mid x,y_{i,<t})}\right)^{\frac{1}{|y_i|}}=\left(\frac{\pi_\theta(y_i\mid x)}{\pi_{\theta_{\mathrm{old}}}(y_i\mid x)}\right)^{\frac{1}{|y_i|}}.
% \end{aligned}
% \]

%\end{align}

% \[
% \mathbf{q}^{l,m}_t\,\mathbf{k}^{l,m\top}_1 \gg \mathbf{q}^{l,m}_t\,\mathbf{k}^{l,m\top}_{j\neq 1}
% \]

% \[
% \cos\!\left(\mathbf{q}^{l,m}_t,\mathbf{k}^{l,h}_1\right)\gg
% \cos\!\left(\mathbf{q}^{l,h}_t,\mathbf{k}^{l,h}_{j\neq 1}\right)
% \]

% \begin{equation}
% \mathbf{o}^{(\ell,m)}_i
% = \sum_{j=1}^{n}\alpha^{(\ell,m)}_{ij}\,\mathbf{v}^{(\ell,m)}_j.
% \label{eq:norm_bound}
% \end{equation}

By the triangle inequality, the output norm is upper bounded by the weighted sum of value norms:
\begin{equation}
\left\|\mathbf{o}^{(\ell,m)}_i\right\|_2
\le \sum_{j=1}^{n}\alpha^{(\ell,m)}_{ij}\,\left\|\mathbf{v}^{(\ell,m)}_j\right\|_2.
\label{eq:norm_bound}
\end{equation}
Therefore, if a substantial portion of attention mass is assigned to tokens with very small
$\left\|\mathbf{v}^{(\ell,m)}_j\right\|_2$, their contribution to $\mathbf{o}^{(\ell,m)}_i$ is correspondingly small.

After concatenating heads and applying the output projection, the multi-head attention update can be written as
\begin{equation}
\mathbf{z}^{(\ell)}_i = \mathbf{W}^{O}\big[\mathbf{o}^{(\ell,1)}_i;\ldots;\mathbf{o}^{(\ell,M)}_i\big],
\label{eq:out_proj}
\end{equation}
and the residual connection updates the hidden state by
\begin{equation}
\mathbf{h}^{(\ell+1)}_i = \mathbf{h}^{(\ell)}_i + \mathbf{z}^{(\ell)}_i.
\label{eq:residual}
\end{equation}
When attending mainly to low-norm value vectors, $\mathbf{z}^{(\ell)}_i$ becomes small in magnitude, making
$\mathbf{h}^{(\ell+1)}_i \approx \mathbf{h}^{(\ell)}_i$.
In this sense, focusing attention on low-norm tokens implements an approximate \emph{no-op} update: the model can
allocate attention probability mass while minimally perturbing representations,
thereby reducing unnecessary information mixing among content-bearing tokens.

\end{document}